# CRIME PREDICTION BASED ON CRIME TYPES AND USING SPATIAL AND TEMPORAL CRIMINAL HOTSPOTS

Tahani Almanie, Rsha Mirza and Elizabeth Lor

Department of Computer Science, University of Colorado, Boulder, USA

*ABSTRACT*

*This paper focuses on finding spatial and temporal criminal hotspots. It analyses two different real-world crimes datasets for Denver, CO and Los Angeles, CA and provides a comparison between the two datasets through a statistical analysis supported by several graphs. Then, it clarifies how we conducted Apriori algorithm to produce interesting frequent patterns for criminal hotspots. In addition, the paper shows how we used Decision Tree classifier and Naïve Bayesian classifier in order to predict potential crime types. To further analyse crimes' datasets, the paper introduces an analysis study by combining our findings of Denver crimes' dataset with its demographics information in order to capture the factors that might affect the safety of neighborhoods. The results of this solution could be used to raise people's awareness regarding the dangerous locations and to help agencies to predict future crimes in a specific location within a particular time.*

## 1. INTRODUCTION

Crimes are a common social problem affecting the quality of life and the economic growth of a society [1]. It is considered an essential factor that determines whether or not people move to a new city and what places should be avoided when they travel [2]. With the increase of crimes, law enforcement agencies are continuing to demand advanced geographic information systems and new data mining approaches to improve crime analytics and better protect their communities [3].

Although crimes could occur everywhere, it is common that criminals work on crime opportunities they face in most familiar areas for them [4]. By providing a data mining approach to determine the most criminal hotspots and find the type, location and time of committed crimes, we hope to raise people's awareness regarding the dangerous locations in certain time periods. Therefore, our proposed solution can potentially help people stay away from the locations at a certain time of the day along with saving lives. In addition, having this kind of knowledge would help people to improve their living place choices. On the other hand, police forces can use this solution to increase the level of crime prediction and prevention. Moreover, this would be useful for police resources allocation. It can help in the distribution of police at most likely crime places for any given time, to grant an efficient usage of police resources [5]. By having all of this information available, we hope to make our community safer for the people living there and also for others who will travel there.

## 2. PROBLEM STATEMENT

Our study aims to find spatial and temporal criminal hotspots using a set of real-world datasets of crimes. We will try to locate the most likely crime locations and their frequent occurrence time. In addition, we will predict what type of crime might occur next in a specific location within a particular time. Finally, we intend to provide an analysis study by combining our findings of a particular crimes dataset with its demographics information.

## 3. RELATED WORK

There has been countless of work done related to crimes. Large datasets have been reviewed, and information such as location and the type of crimes have been extracted to help people follow law enforcements. Existing methods have used these databases to identify crime hotspots based on locations. There are several maps applications that show the exact crime location along with the crime type for any given city (see Figure 1). Even though crime locations have been identified, there is no information available that includes the crime occurrence date and time along with techniques that can accurately predict what crimes will occur in the future.

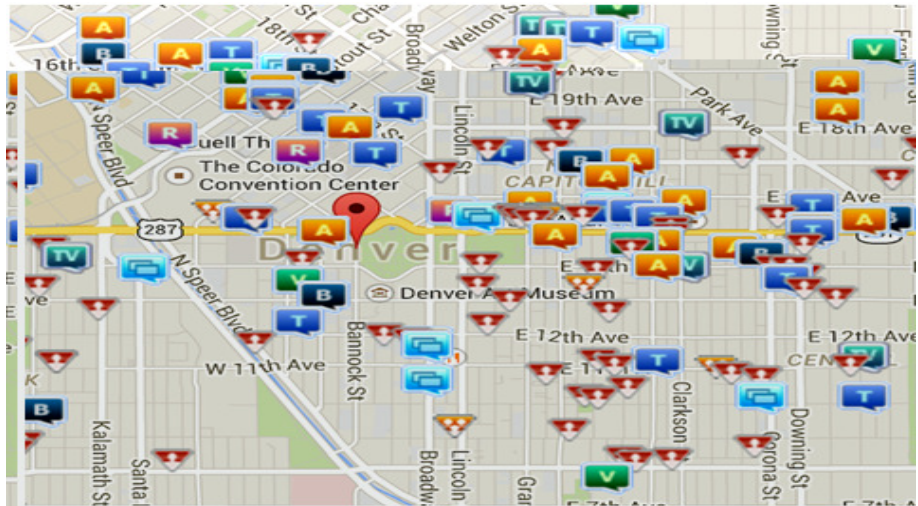
Figure 1. Map of downtown Denver, Colorado with crime locations [6]

On the other hand, the previous related work and their existing methods mainly identify crime hotspots based on the location of high crime density without considering either the crime type or the crime occurrence date and time. For example, related research work containing a dataset for the city of Philadelphia with crime information from year 1991 - 1999. It was focusing on the existence of multi-scale complex relationships between both space and time [1]. Another research titled "The utility of hotspot mapping for predicting spatial patterns of crime" looks at the different crime types to see if they differ in their prediction abilities [7]. Other existing works explore relationships between the criminal activity and the socio-economics variables such as education, ethnicity, income level, and unemployment [1].

International Journal of Data Mining & Knowledge Management Process (IJDKP) Vol.5, No.4, July 2015

Despite all of the existing work, none of them consider the three elements (location, time, crime type) together. In addition, there is very little research that can accurately predict where crimes will happen in the future [7]. In our study, we provide a data-mining model for crime prediction based on crime types and using spatial and temporal criminal hotspots.

## 4. DATASETS

In our study, we used two different datasets for real-word crimes in two cities of the US. We chose those cities from different states: Denver in Colorado, Los Angeles in California. To construct our data mining models, we mainly focused on Denver dataset. After we had built the desired models, we applied the same strategy to train the required models on Los Angeles dataset. Additionally, we combined our mining findings of Denver crimes dataset with its demographics information using another dataset of Denver Neighborhood Demographics. In this section, we give a description for our three different datasets.

### 4.1. Denver Crimes Dataset

This dataset represents the real-world crimes in Denver, Colorado. It includes criminal offenses and crime incidents in the city and county of the city for the previous five calendar years in addition to the current year (2010 – 2015). The dataset information is based on the National Incident Based Reporting System (NIBRS) [8]. This dataset is composed of 19 attributes with 333068 instances. The key attributes provide the offense type and its category such as robbery, public-disorder, and sexual assault. The dataset also gives the exact occurrence time of the crime along with the district, the neighborhood and the exact geographic location. The following table shows the used key attributes and its content values (Table 1).

Table 1. Denver key attributes table

| Attribute | Data Type | Number of Distinct Values | Value |
|---|---|---|---|
| Offense_Category_Id | Nominal | 14 Categories | Aggravated-assault<br>All-other-crimes<br>Arson<br>Burglary<br>Drug-alchol<br>Larceny<br>Murder<br>Other-crimes-against-persons<br>Public-disorder<br>Robbery<br>Sexual assault<br>Theft-from-motor-vehicle<br>White-collar-crime |
| First_Occurance_Date | Date and time | Unlimited | 6/13/14 21:30 |
| Neighborhood_Id | Nominal | 78 Neighborhoods | (See Figure 2) |
| Is_Crime | Binary | 2 values | 0 or 1 |

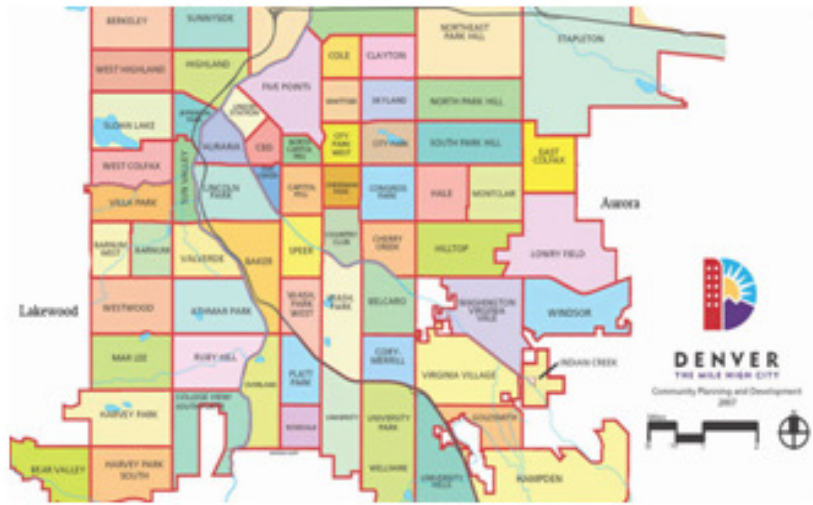

Figure 2. Map of Denver neighborhoods [9]

### 4.2. Los Angeles Crimes Dataset

This dataset represents the real-world crimes in Los Angeles, California. It includes criminal offenses and crime incidents in the city and the area of the city. 96% of the crimes in the dataset occurred in the year 2014 while the other 4% of the crimes occurring before 2014. This dataset information was obtained from the US City Open Data Census [10]. It is composed of 14 attributes with 243750 instances. Unlike the Denver dataset, the crime category is more specific with its crime such as Theft-Person, Theft-Plain, and Theft-From-Motor-Vehicle. The following table shows the used key attributes and its content values (Table 2).

Table 2. Los Angeles key attributes table

| Attribute | Data Type | Number of Distinct Values | Value |
|---|---|---|---|
| Crm Cd Desc | Nominal | 128 Categories | Burglary Robbery Vandalism Aggravated Assault etc. |
| Date Occ | Date | Unlimited | 8/23/14 |
| Time Occ | Time | Unlimited | 2200 |
| Area Name | Nominal | 21 Names | (See Figure 3) |

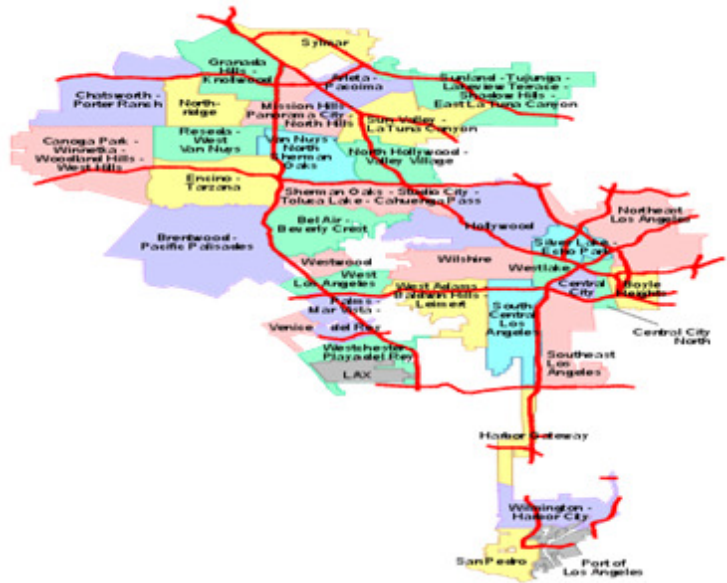

Figure 3. Map of Los Angeles areas [11]

### 4.3. Denver Neighborhood Demographics Dataset

This dataset represents the population and housing data of Denver city and its county. The information was combined from Community Planning and Development along with 2010 US Census [12]. The dataset has 127 attributes and 78 instances. It provided detailed information about people living in each of the 78 neighborhoods in Denver. Some of the attributes that we were interested in for our mining analysis are gender, race, age, family size, housing units, number of occupied and vacant units, and number of rental and owned units.

## 5. METHODOLOGY

We strongly believe that finding relationships between crime elements could highly help in predicting potential dangerous hotspots at a certain time in the future. Therefore, our proposed approach aimed to focus on three main elements of crimes data, which are the type of crime, the occurrence time and the crime location. We tried to extract all possible interesting frequent patterns based on the crime variables. Then, we applied some classification methods in order to predict potential crime types in a specific location within a particular time.

In this section, we explain how we prepared our datasets. After that, we provide how we analysed the data using some statistical analysis. Then, we introduce how we constructed our data-mining models to achieve our purpose.

### 5.1. Data Preprocessing

We performed the following preprocessing steps on the two datasets:

**5.1.1. Data Cleaning**

There are some missing values in some attributes such as last_occurance_date and incident_address in Denver dataset. However, we found that all attributes containing missing values are not of our key attributes. Therefore, we did not need to clean them. All key attributes in (Table 1 and Table 2) were completed with cleaned values in both datasets. In addition, we did not found any noisy or inconsistent values in these attributes.

**5.1.2. Data Reduction**

For both crime datasets, we needed to apply data reduction. We implemented dimensionality reduction using attribute subset selection. For example, among the available 19 attributes in Denver crimes dataset, we just selected four of them. The selected attributes are the related ones or the key attributes for our mining purpose (see Table 1). We removed all the other irrelevant attributes from the dataset.

On the other hand, we performed data reduction in terms of number of instances. We observed that Denver crimes dataset contained a set of traffic accident instances. The attribute "Is_Crime" indicates whether the instance belongs to a crime or accident. While we concern with crime information, we used the attribute "Is_Crime" to filter the instances and remove all the irrelevant ones. We applied the same strategy for Los Angeles crimes dataset. After reduction, we ended up with having 231640 instances in Denver and 196767 instances in Los Angeles.

**5.1.3. Data Integration**

We performed several steps of data integration for our datasets. First, to avoid different attribute naming, we unified the key attribute names for both crime datasets as follow: Crime_Type, Crime_Date, and Crime_Location. Crime_Location represents the neighborhood attribute for Denver dataset whereas the Area attribute for Los Angeles dataset. Our mining study requires analyzing the date and time info on different granularities. Therefore, we used the Crime_Date attribute, which contains date and time crime info, to generate three more attributes: Crime_Month, Crime_Day, and Crime_Time. We adopted the military time system, and we considered the hour part without paying attention to the minutes to get more of frequent patterns. In addition, we initiated Crime_Type_Id attribute to give an id for each of the 14 crime categories (See Table1). We used this attribute for both datasets to get integrated crime types.

**5.1.4. Data Transformation and Discretization**

We finished our data integration process by having 24 different distinct values for the Crime_Time attribute and 14 types for the Crime_Type attribute. We realized that it is necessary to reduce the diversity of these two attribute values. Thus, we applied data transformation to both attributes by mapping their values to fall within smaller groups. Our goal was to get more frequent patterns and to increase the model accuracy. For the crime types feature, we minimize the type list by grouping them into six new types. For the crime time feature, we mapped its values into 4-hour intervals. Table 3 illustrates the resulted attributes after data preprocessing.

Table 3. Resulted attributes after data preprocessing

| Attribute | Number of Distinct Values | Value |
|---|---|---|
| Crime_Type (nominal) | 6 | Assault<br>Drug Alcohol<br>Other crimes<br>Public Disorder<br>Theft<br>White collar crime |
| Crime_Type_Id (numeric) | 6 | 1: Assault<br>2: Drug Alcohol<br>3: Other crimes<br>4: Public Disorder<br>5: Theft<br>6: White collar crime |
| Crime_Month (nominal) | 12 | months names |
| Crime_Day (nominal) | 7 | days of the week |
| Crime_Time (nominal) | 6 | T1: 1am to 4:59 am<br>T2: 5 am to 8:59 am<br>T3: 9 am to 12:59 pm<br>T4: 13 pm to 16:59 pm<br>T5: 17 pm to 20:59 pm<br>T6: 21 pm to 0:59 am |
| Crime_Location (nominal) | Denver: 78<br>LA: 21 | (See Figure 2)<br>(See Figure 3) |

## 5.2. Data Analysis

As an initial step to analyse and get the big view of our data, we conducted statistical analysis on the attribute values of our datasets. For each city, we started with generating a python script to calculate frequencies of distinct values for every attribute. Then, we created a variety of graphs to give us better understanding of our data. Each graph came up with the percentage of crime occurrences regarding a particular aspect.

Figures 4 – 6, provide a statistical comparison between Denver and Los Angeles crime datasets. While most of crimes in Los Angeles dataset belong to 2014, we chose to limit the comparison for crimes committed in the year 2014 for both cities in order to get a fair comparison. Also, we graphed over the percentage of crime occurrence instead of the total number of crimes committed. Using the percentage can help to better understand and visualize the differences and similarities between the two cities.

Figure 4 shows the percentage of crime occurrences over the 12 months in Denver and Los Angeles. While the month of August has the maximum number of crimes in Denver, in Los Angeles, most of the crimes happen in the month of July. On the other hand, November has the least number of crimes in Denver whereas February appeared to be the safest month in Los Angeles. Overall, there are some similarities between the two cities for the crime distribution over the 12 months.

Figure 5 shows the percentage of crime occurrences over the days of the week in Denver and Los Angeles. By looking at the graph, it is obvious that Friday is the peak day of crimes for both Denver and Los Angeles. On the other hand, Sunday has the lowest crime rate for both cities.

Figure 6 shows the percentage of crime occurrences over the 24 hours in Denver and Los Angeles. In Denver around 25% of crimes happen in the late evening hours starting from 9pm to 1am. For Los Angeles, from 5pm to 9pm is the peak time period at which most crimes occur. The early morning hours seem to be safer than the rest of day for both cities.

Figure 7 and Figure 8 show the percentage of all crime occurrences over different locations in Denver and Los Angeles. With many locations for both cities, we chose to display ten different neighborhoods\areas for both cities. We graphed the locations that represented the top 3, middle 4 and least 3 locations in terms of their crime rates. For Denver, the graph shows the neighborhood, Five Points having the most crime rate among the 78 neighborhoods while Wellshire has the minimum crime rate with less than 0.2 % of the total crimes. For Los Angeles, the district, 77th Street, seems to be the most dangerous area whereas Hollenbeck appears to be relatively the safest one.

Figures 9 – 12, represented the crime statistics for Denver and Los Angeles taking into account the different crime types. This helped to give a better overall summary of the different crimes types that happen throughout the days of the week and along the 24 hours in both cities. It is clear from both cities' graphs that the most common crime type is Theft, and it is evenly distributed along the days of the week.

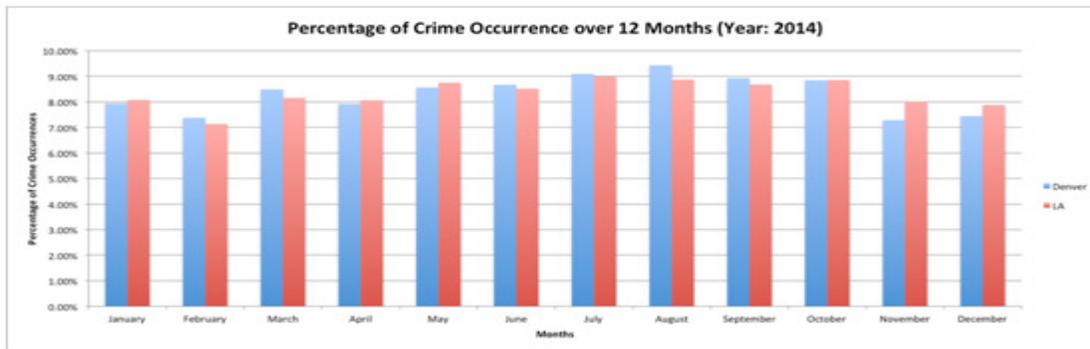

Figure 4. Crime rate over the 12 months in Denver and Los Angeles in 2014

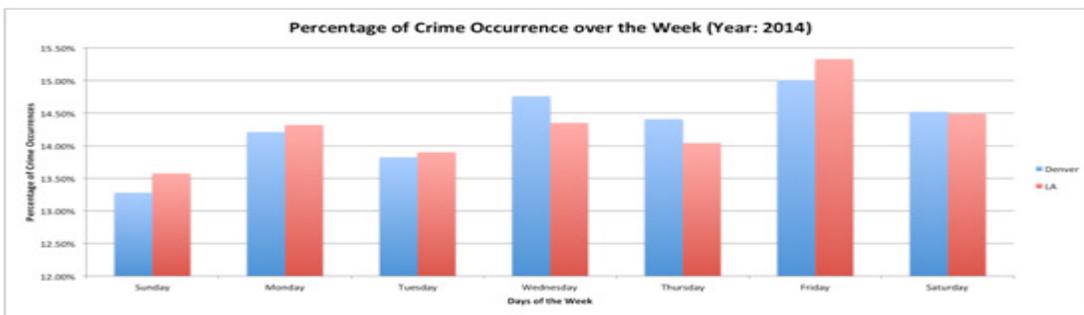

Figure 5. Crime rate over the days of the week in Denver and Los Angeles in 2014

International Journal of Data Mining & Knowledge Management Process (IJDKP) Vol.5, No.4, July 2015

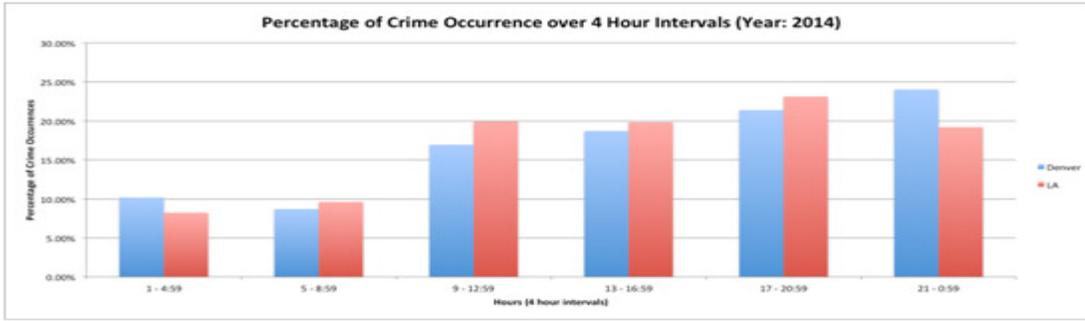

Figure 6. Crime rate over the 24 hours in Denver and Los Angeles in 2014

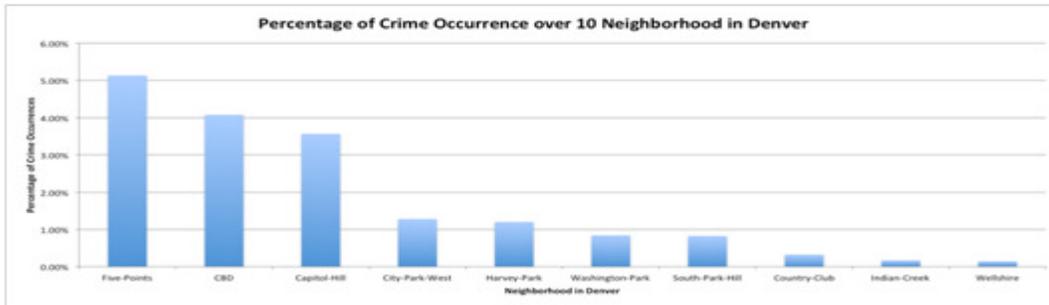

Figure 7. Crime rate over 10 neighborhoods in Denver

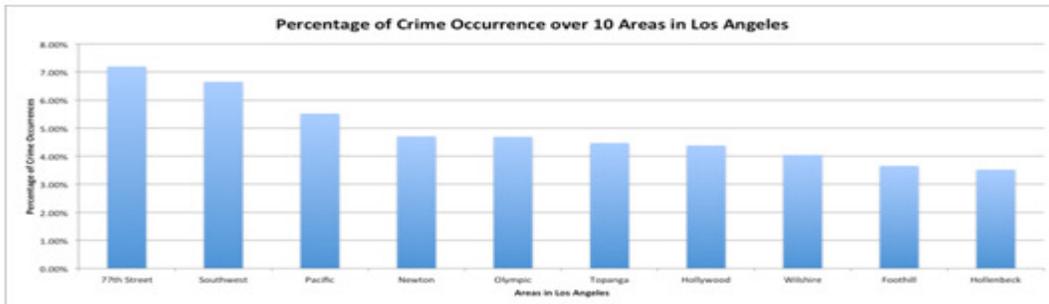

Figure 8. Crime rate over 10 areas in Los Angeles

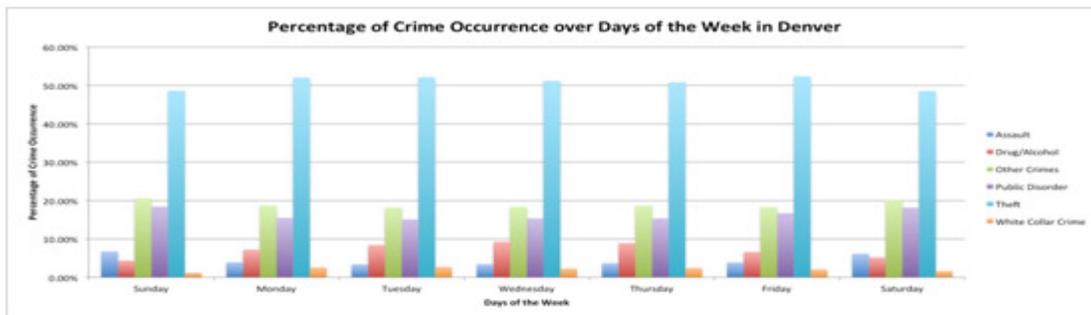

Figure 9. Percentage of crimes over the days of the week in Denver based on the different crime types

International Journal of Data Mining & Knowledge Management Process (IJDKP) Vol.5, No.4, July 2015

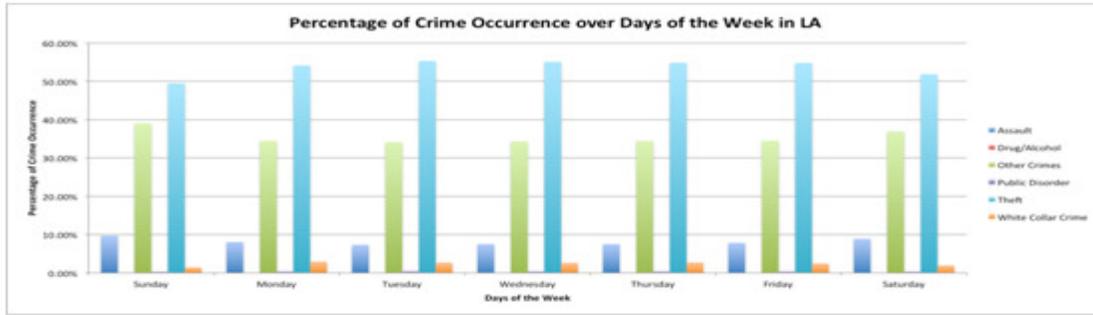

Figure 10. Percentage of crimes over the days in LA based on the different crime types

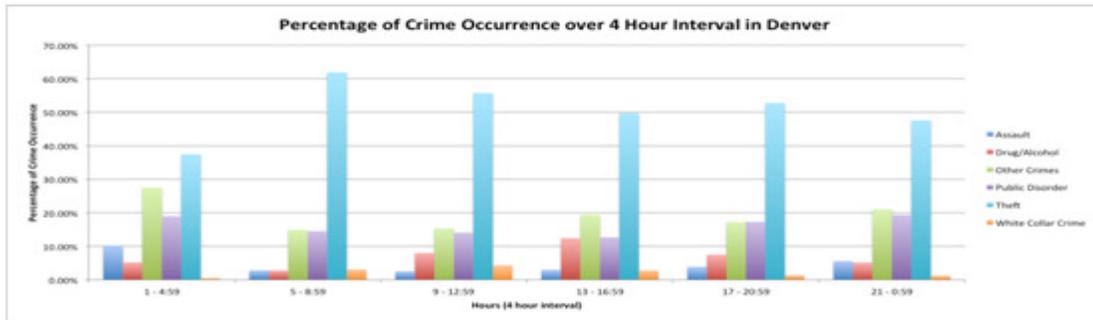

Figure 11. Percentage of crimes over the 24 hours in Denver based on the different crime types

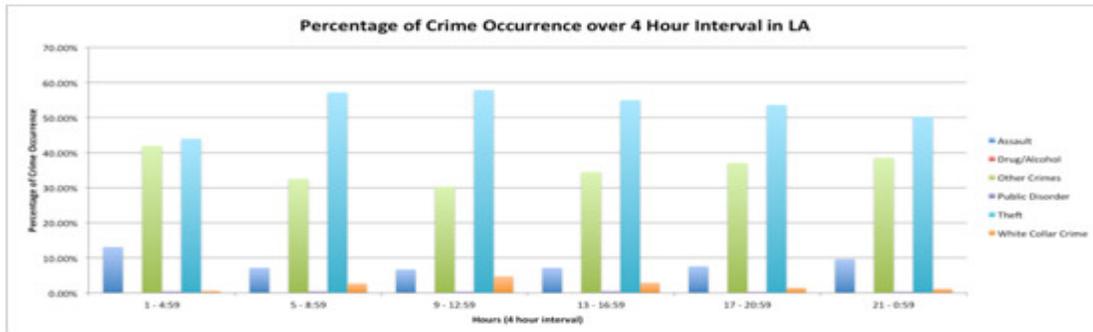

Figure 12. Percentage of crimes over the 24 hours in LA based on the different crime type

## 5.3. Models Building

In order to extract frequent patterns from Denver and Los Angeles crimes, we applied the Apriori algorithm on both datasets. Then we use Naïve Bayesian classifier and decision tree classifier to build two different classification models for each dataset. The purpose of the classifiers is to predict the potential crime type in a specific location within a particular time in the future. We aimed to examine every model then choose the model that gives the best accuracy in prediction. In this section, we provide a brief description of each model used.

### 5.3.1. Apriori Algorithm

Apriori is one of the basic algorithms for mining frequent patterns. It scans the dataset to collect all itemsets that satisfy a predefined minimum support. Our goal of using this model is to find all possible crime frequent patterns regardless of the committed crime type. We wanted to come up with a list of all crime hotspots along with its related frequent time. Hence, we implemented the algorithm on location and time features and excluded the crime type feature. Additionally, to obtain more frequent patterns we applied constraint-based mining by restricting the extraction process on the frequent patterns having this formula of three specific itemsets (Location, Day, Time).

We implemented this model using an open source tool [13]. We conducted multiple experiments using different minimum support values for each dataset. Then we selected the optimum choice. For Denver, the minimum support value was 0.0012, which corresponds to 277 absolute frequencies. For Los Angeles, the minimum support value was 0.0018, which corresponds to 354 absolute frequencies.

### 5.3.2. Naïve Bayesian Classifier

Naïve Bayesian classifier is a supervised learning algorithm, which is effective and widely used. It is a statistical model that predicts class membership probabilities based on Bayes' theorem (Formula 3). It assumes the independent effect between attribute values. While our selected crime features have an independent effect on each other, this classifier was an ideal choice.

$$P(H|X) = P(X|H) P(H) / P(X) \qquad (3)$$

We constructed this model using Scikit–Learn that provides a set of open source data-mining tools for Python. We applied Multinomial Naïve Bayes, which is used for multinomial distributed data that conforms to the categorical features in our datasets. The crime features contain (month, day, time, location) of the crime while we selected the crime type to represent the class label. We randomly divided the dataset into 80% of data as a training set and 20% of data as a testing set. We trained the same classifier on the training data for each of Denver and Los Angeles datasets to obtain two different models ready for crime type prediction in each of the two cities.

### 5.3.3. Decision Tree Classifier

Decision Tree classifier is our second used supervised learning algorithm. It creates a model to predict the class label values by learning simple decision rules implied from the data features.

We created this model for both datasets using Scikit–Learn another library tool allocated for decision tree induction. To measure the quality of the split, we applied the entropy function for the information gain. Figure 14 shows a partial view of the constructed tree on Los Angeles training dataset. Since the generated tree was complex, we restricted the decision tree to have ten maximum leaf nodes. The tree shows that the Time attribute is selected as the root node to split the data.

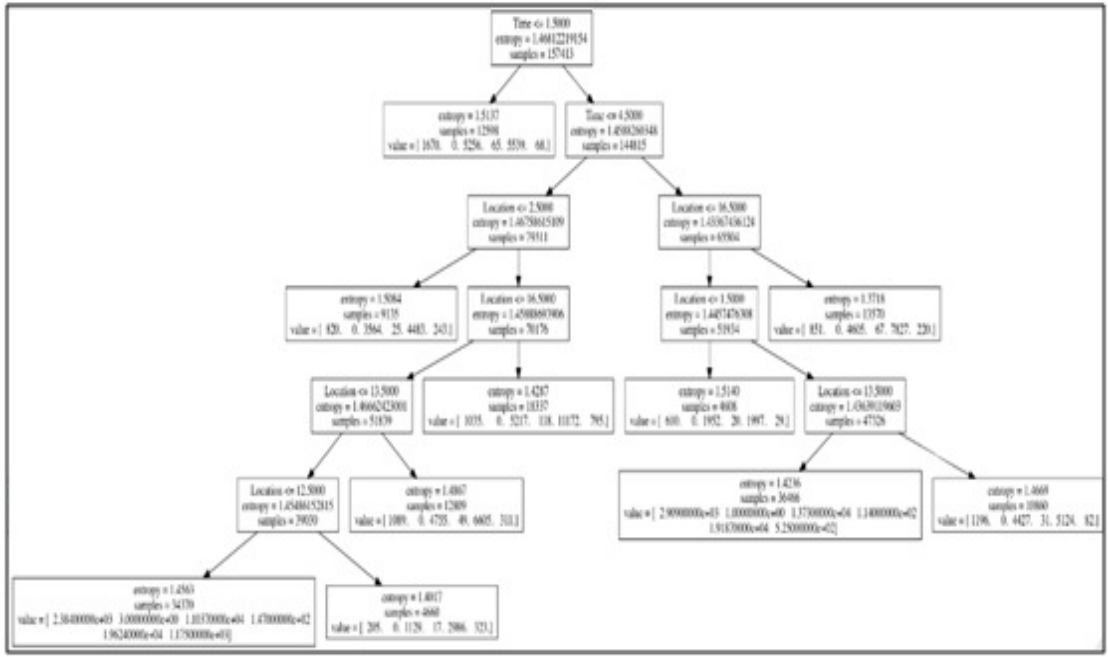

Figure 14. Partial view of the applied decision tree on Los Angeles training dataset

## 6. EVALUATION

In this section, we evaluate each of the three constructed models regarding different aspects.
Apriori Algorithm is our first model that was used to extract frequent crime patterns. The key strength for this model are its readiness and easiness of use and implement. On the contrary, the main drawback is slowness. It takes very long running time to give the results, especially with smaller values of the minimum support.

The other two models are Naïve Bayesian and decision tree classifiers that were used for crime type prediction. We applied the 5-fold cross validation strategy on both models then compared the prediction accuracy for each city. Regarding to the Naïve Bayesian classifier, it achieves an accuracy of 51% in Denver crime prediction while it reaches 54% for Los Angeles crime prediction. On the other hand, decision tree classifier reports less prediction accuracy with 42 % for Denver and 43% for Los Angeles. Moreover, the decision tree model created a very complex tree that cannot generalize the data for both cities. However, the two classifiers have the same performance in terms of their running time.

Since the Bayesian classifier yields the best overall performance, we chose it as the ideal model for crime prediction in our study. Table 4 reports the confusion matrix of applying this model on Denver testing set. Table 5 shows a report of the main classification metrics obtained by this model.

Table 4. Confusion matrix of Bayesian classifier

| Actual class\ Predicted | Assault | Drug alcohol | Other crimes | Public disorder | Theft | White collar crime |
|---|---|---|---|---|---|---|
| Assault | 0 | 4 | 230 | 0 | 1833 | 0 |
| Drug alcohol | 0 | 39 | 344 | 0 | 3004 | 0 |
| Other crimes | 0 | 42 | 1028 | 0 | 77738 | 0 |
| Public disorder | 0 | 10 | 409 | 0 | 7159 | 0 |
| Theft | 0 | 27 | 737 | 0 | 22721 | 0 |
| White collar crime | 0 | 2 | 30 | 0 | 971 | 0 |

Table 5. Classification report of Bayesian classifier

|  | Precision | Recall | F1-score | Support |
|---|---|---|---|---|
| Assault | 0.00 | 0.00 | 0.00 | 2067 |
| Drug alcohol | 0.31 | 0.01 | 0.02 | 3387 |
| Other crimes | 0.37 | 0.12 | 0.18 | 8808 |
| Public disorder | 0.00 | 0.00 | 0.00 | 7578 |
| Theft | 0.52 | 0.97 | 0.68 | 23485 |
| White collar crime | 0.00 | 0.00 | 0.00 | 1003 |
| Avg/Total | 0.36 | 0.51 | 0.38 | 46328 |

The precision is the ratio tp / (tp + fp) where tp is the number of true positives and fp the number of false positives. The recall is the ratio tp / (tp + fn) where tp is the number of true positives and fn the number of false negatives. F1 score is a weighted average of the precision and recall [14].

## 7. KEY RESULTS

In this section, we summarize the key results that we obtained from applying the Apriori and Bayesian classifier models on the two datasets. Then, we provide an analysis study through combining our findings of Denver crimes dataset with its demographics information.

### 7.1. Crime Frequent Hotspots

The first goal of our study was finding spatial and temporal criminal hotspots. We have successfully achieved this goal using Apriori algorithm on both Denver and Los Angeles datasets. We have extracted all the interesting patterns based on our predefined thresholds. We found that Denver has 62 interesting frequent patterns while Los Angeles has 59 patterns. Table 4 and Table 5 report our Apriori results for Denver and Los Angeles crime frequent patterns. The frequent itemsets ordered by the location, day of the week, and the time period. With these different frequent itemsets, we are able to conclude the most likely crime locations along with their frequent occurrence day and time.

Table 6 indicates that Five-Point, Capitol Hill, CBD, Montebello, Union Station, Stapleton, and Westwood are the hotspots that have most crimes frequent patterns in Denver. It is obvious that Five-Point has the largest number of patterns compared to other locations while CBD comes next. In addition, we can find that Wednesday is the peak day of crimes occurred in CBD. It is also interesting to notice that Union Station has frequent patterns only on weekend days (Friday, Saturday, Sunday) four hours before and after midnight.

In Los Angeles, we can see that most likely crimes happen at 77th Street, Southwest, Pacific, N Hollywood, Southeast, Northeast, and Van Nuys respectively. Among all crime patterns, the highest frequent one occurs in 77th Street on Monday around 9pm to 1am. Both 77th Street and Southwest areas have crimes everyday, and their crimes are more likely to happen from 8 am to midnight (Table 7).

## 7.2. Crime Prediction

The second target for our study was to predict the crime type that might occur in a specific location within a particular time. The Bayesian classifier enabled us to reach this target with a reasonable accuracy. To predict an expected crime type, you need to provide four related features of the crime. The required features are: the occurrence month, the occurrence day of the week, the occurrence time and the crime location. All features can be submitted in their nominal values. The provided occurrence time should be in the form of time period interval from T1 to T6 (See Table 3). For Denver, the location has to be one of its 78 neighborhoods (See Figure 2). For Los Angeles, the location should be one of its 21 areas (See Figure 3). Every given result is a number from 1 to 6 that indicates the predicted crime type for a given set of crime features. Table 3 gives the corresponding crime type for each number.

Table 6. Apriori results for Denver crime frequent patterns

| Frequent pattern | Min-sup | Frequent pattern | Min-sup |
|---|---|---|---|
| 'Capitol-hill', 'Monday', 'T5' | 0.001 | 'Five-points', 'Thursday', 'T4' | 0.001 |
| 'Capitol-hill', 'Thursday', 'T6' | 0.001 | 'Five-points', 'Thursday', 'T5' | 0.002 |
| 'Capitol-hill', 'Friday', 'T5' | 0.001 | 'Five-points', 'Thursday', 'T6' | 0.002 |
| 'Capitol-hill', 'Friday', 'T6' | 0.002 | 'Five-points', 'Wednesday', 'T3' | 0.001 |
| 'Capitol-hill', 'Saturday', 'T6' | 0.002 | 'Five-points', 'Wednesday', 'T4' | 0.002 |
| 'Capitol-hill', 'Sunday', 'T6' | 0.001 | 'Five-points', 'Wednesday', 'T5' | 0.002 |
| 'CBD', 'Monday', 'T4' | 0.001 | 'Five-points', 'Wednesday', 'T6' | 0.002 |
| ' CBD ', 'Monday', 'T5' | 0.001 | 'Five-points', 'Saturday', 'T1' | 0.001 |
| ' CBD ', 'Tuesday', 'T3' | 0.001 | 'Five-points', 'Saturday', 'T5' | 0.002 |
| ' CBD ', 'Tuesday', 'T4' | 0.001 | 'Five-points', 'Saturday', 'T6' | 0.002 |
| ' CBD ', 'Wednesday', 'T3' | 0.001 | Five-points', 'Sunday', T1' | 0.001 |
| ' CBD ', 'Wednesday', 'T4' | 0.002 | 'Five-points', 'Sunday', 'T5' | 0.001 |
| ' CBD ', 'Wednesday', 'T5' | 0.001 | 'Five-points', 'Sunday', 'T6' | 0.002 |
| ' CBD ', 'Wednesday', 'T6' | 0.001 | 'Montebello', 'Monday', 'T6' | 0.001 |
| ' CBD ', 'Thursday', 'T3' | 0.001 | 'Montebello', 'Wednesday', 'T6' | 0.001 |
| ' CBD ', 'Thursday', 'T4' | 0.001 | 'Montebello', 'Thursday', 'T6' | 0.001 |
| ' CBD ', 'Thursday', 'T5' | 0.002 | 'Montebello', 'Friday', 'T5' | 0.001 |
| ' CBD ', 'Friday', 'T4' | 0.001 | 'Montebello', 'Friday', 'T6' | 0.001 |
| ' CBD ', 'Friday', 'T5' | 0.001 | 'Montebello', 'Saturday', 'T6' | 0.002 |
| ' CBD ', 'Friday', 'T6' | 0.001 | 'Montebello', 'Sunday', 'T6' | 0.002 |
| ' CBD ', 'Saturday', 'T5' | 0.002 | 'Stapleton', 'Wednesday', 'T5' | 0.001 |
| ' CBD ', 'Saturday', 'T6' | 0.002 | 'Stapleton', 'Friday', 'T5' | 0.002 |
| 'Five-points', 'Monday', 'T5" | 0.002 | 'Union-station', 'Friday', 'T6' | 0.001 |
| | | | 0.002 |
| 'Five-points', 'Monday', 'T6' | 0.002 | 'Union-station', 'Saturday', 'T1' | 0.002 |
| 'Five-points', 'Tuesday', 'T3' | 0.001 | 'Union-station', 'Saturday', 'T6' | 0.002 |
| 'Five-points', 'Tuesday', 'T4' | 0.001 | 'Union-station', 'Sunday', 'T1' | 0.001 |
| 'Five-points', 'Tuesday', 'T5' | 0.002 | 'Union-station', 'Sunday', 'T6' | 0.001 |
| 'Five-points', 'Tuesday', 'T6' | 0.001 | 'Westwood', 'Thursday', 'T6' | 0.001 |
| 'Five-points', 'Friday', 'T5' | 0.002 | 'Westwood', 'Friday', 'T6' | 0.002 |
| 'Five-points', 'Friday', 'T6' | 0.002 | 'Westwood', 'Saturday', 'T6' | 0.001 |
| 'Five-points', 'Thursday', 'T3' | 0.001 | 'Westwood', 'Sunday', 'T6' | |

Table 7. Apriori results for Los Angeles crime frequent patterns

| Frequent pattern | Min-sup | Frequent pattern | Min-sup |
|---|---|---|---|
| 'N Hollywood', 'Tuesday', 'T5' | 0.002 | 'Southwest', 'Saturday', 'T4' | 0.002 |
| 'N Hollywood', 'Thursday', 'T3' | 0.002 | 'Southwest', 'Saturday', 'T5' | 0.002 |
| 'N Hollywood', 'Friday', 'T5' | 0.002 | 'Southwest', 'Sunday', 'T4' | 0.002 |
| 'N Hollywood', 'Saturday', 'T5' | 0.002 | 'Southwest', 'Sunday', 'T5' | 0.002 |
| 'Northeast', 'Friday', 'T5' | 0.002 | 'Van Nuys', 'Friday', 'T5' | 0.002 |
| 'Pacific', 'Monday', 'T3' | 0.002 | '77th St', 'Monday', 'T3' | 0.002 |
| 'Pacific', 'Tuesday', 'T3' | 0.002 | '77th St', 'Monday', 'T4' | 0.002 |
| 'Pacific', 'Wednesday', 'T3' | 0.002 | '77th St', 'Monday', 'T5' | 0.003 |
| 'Pacific', 'Thursday', 'T3' | 0.002 | '77th St', 'Monday', 'T6' | 0.002 |
| 'Pacific', 'Friday', 'T3' | 0.002 | '77th St', 'Tuesday', 'T3' | 0.002 |
| 'Southeast', 'Monday', 'T5' | 0.002 | '77th St', 'Tuesday', 'T4' | 0.002 |
| 'Southeast', 'Thursday', 'T5' | 0.002 | '77th St', 'Tuesday', 'T5' | 0.002 |
| 'Southeast', 'Friday', 'T5' | 0.002 | '77th St', 'Wednesday', 'T3' | 0.002 |
| 'Southeast', 'Sunday', 'T5' | 0.002 | '77th St', 'Wednesday', 'T4' | 0.002 |
| 'Southwest', 'Monday', 'T3' | 0.002 | '77th St', 'Wednesday', 'T5' | 0.002 |
| 'Southwest', 'Monday', 'T4' | 0.002 | '77th St', 'Wednesday', 'T6' | 0.002 |
| 'Southwest', 'Monday', 'T5' | 0.002 | '77th St', 'Thursday', 'T3' | 0.002 |
| 'Southwest', 'Tuesday', 'T3' | 0.002 | '77th St', 'Thursday', 'T4' | 0.002 |
| 'Southwest', 'Tuesday', 'T4' | 0.002 | '77th St', 'Thursday', 'T5' | 0.002 |
| 'Southwest', 'Tuesday', 'T5' | 0.002 | '77th St', 'Friday', 'T3' | 0.002 |
| 'Southwest', 'Wednesday', 'T3' | 0.002 | '77th St', 'Friday', 'T4' | 0.002 |
| 'Southwest', 'Wednesday', 'T4' | 0.002 | '77th St', 'Friday', 'T5' | 0.002 |
| 'Southwest', 'Wednesday', 'T5' | 0.002 | '77th St', 'Friday', 'T6' | 0.002 |
| 'Southwest', 'Friday', 'T3' | 0.002 | '77th St', 'Saturday', 'T3' | 0.002 |
| 'Southwest', 'Friday', 'T4' | 0.002 | '77th St', 'Saturday', 'T5' | 0.002 |
| 'Southwest', 'Friday', 'T5' | 0.002 | '77th St, 'Saturday', 'T6' | 0.002 |
| 'Southwest', 'Thursday', 'T3' | 0.002 | '77th St', 'Sunday', 'T3' | 0.002 |
| 'Southwest', 'Thursday', 'T4' | 0.002 | '77th St', 'Sunday', 'T5' | 0.002 |
| 'Southwest', 'Thursday', 'T5' | 0.002 | '77th St', 'Sunday', 'T6' | 0.002 |
| 'Southwest', 'Saturday', 'T3' | 0.002 | | |

### 7.3. Crime Hotspots Demographics Analysis

After accomplishing our main goal by locating spatial and temporal criminal hotspots and predicting potential crime types, we applied some demographics analysis using Denver neighborhood demographics dataset. We wanted to further understand our models' findings by studying the relationship between crime rate in each neighborhood and its demographics

International Journal of Data Mining & Knowledge Management Process (IJDKP) Vol.5, No.4, July 2015information. In other words, we wanted to find factors that affect the safety of Denver neighborhoods. Hence, we performed a demographics comparison between the most three dangerous neighborhoods (Five Points, CBD, and Capitol Hill) with the three safest neighborhoods in Denver (Wellshire, Indian Creek, and Country Club) to find out the main demographics characteristics of each group.

By studying the demographics data for each neighborhood, we found that dangerous neighborhoods in Denver associated with large population and large number of housing units (Figures 15 and 16). Additionally, there is an interesting correlation between the big numbers of vacant houses and the dangerous locations (Figure 17).

Moreover, we found that people's age and gender distribution vary between dangerous and safe locations. Specifically, dangerous neighborhoods have more male while the safe neighborhoods have more female (Figure 18). Furthermore, dangerous neighborhoods tend to have bigger number of people with ages from 20 to 29 years whereas the safest neighborhoods seem to include larger number of people with ages from 50 to 59 years (Figure 19). However, we have not found any relationship between crime hotspots and people's race distribution.

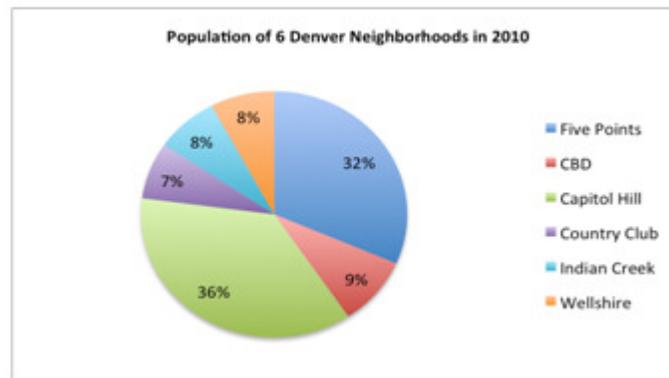

Figure 15. Population of six neighborhoods of Denver in 2010

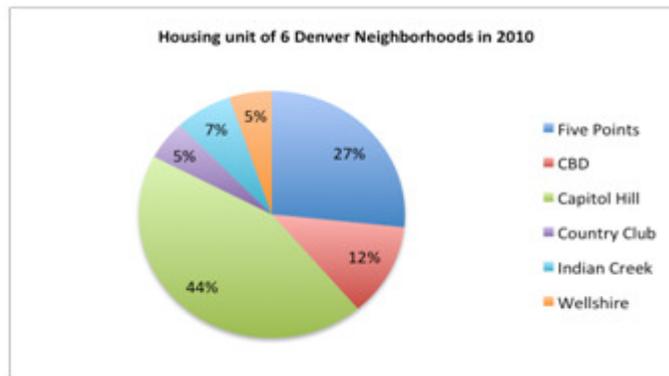

Figure 16. Total housing units in six neighborhoods of Denver in 2010

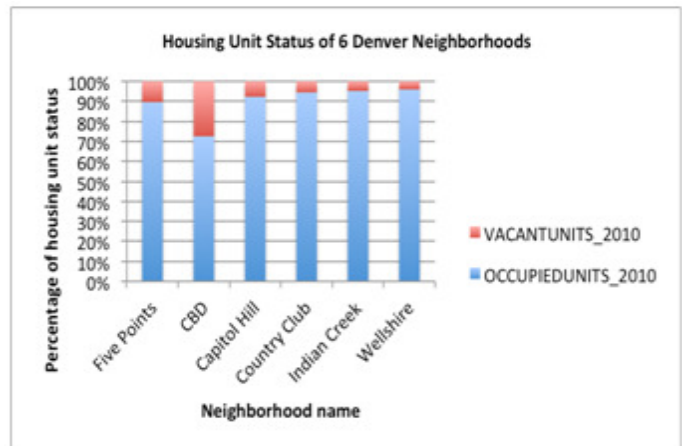

Figure 17. Housing units' status in six neighborhoods of Denver

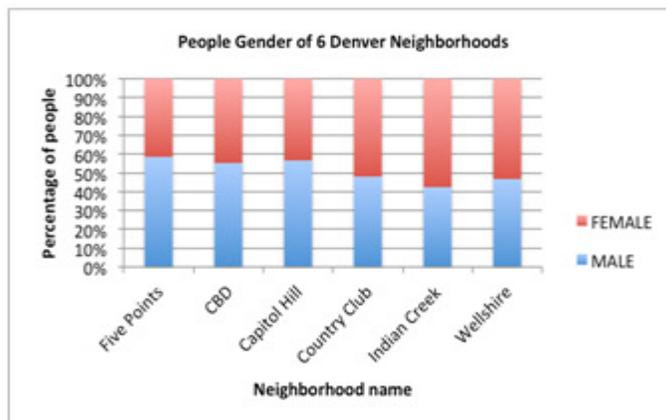

Figure 18. People gender in six neighborhoods of Denver

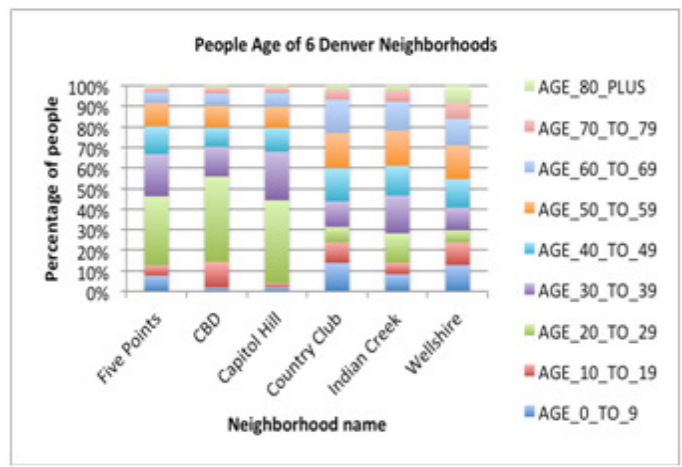

Figure 19. People age distribution in six neighborhoods of Denver

## 8. CONCLUSIONS AND FUTURE WORK

We generated many graphs and found interesting statistics that showed the baseline to understand Denver and Los Angeles crimes datasets. Then, we applied Apriori algorithm to find frequent crime patterns in both cities. After that, we applied Decision Tree and Naïve Bayesian classifiers to help predicting future crimes in a specific location within a particular time. We achieved 51% of prediction accuracy in Denver and 54% prediction accuracy in Los Angeles. Finally, we provided an analysis study by combining our findings of Denver crimes' dataset with its demographics information. We aimed to further understand our models' findings and to capture the factors that might affect the safety of neighborhoods.

As a future extension of our work, we plan to apply more classification models to increase crime prediction accuracy and to enhance the overall performance. It is also a helpful extension for our study to consider the income information for neighborhoods in order to see if there are relationships between neighborhoods income level and their crime rate. Moreover, we intend to analyse Los Angeles demographics information with its crime findings. Furthermore, we want to study other crimes datasets from new cities along with their demographics datasets.

Last but not least, we hope by publishing this paper starting a trend of crimes prediction, which can help law enforcements and keep our community safer for everyone.

## REFERENCES

[1] A. Bogomolov, B. Lepri, J. Staiano, N. Oliver, F. Pianesi and A. Pentland, 'Once Upon a Crime: Towards Crime Prediction from Demographics and Mobile Data', CoRR, vol. 14092983, 2014.
[2] R. Arulanandam, B. Savarimuthu and M. Purvis, 'Extracting Crime Information from Online Newspaper Articles', in Proceedings of the Second Australasian Web Conference - Volume 155, Auckland, New Zealand, 2014, pp. 31-38.
[3] A. Buczak and C. Gifford, 'Fuzzy association rule mining for community crime pattern discovery', in ACM SIGKDD Workshop on Intelligence and Security Informatics, Washington, D.C., 2010, pp. 1-10.
[4] M. Tayebi, F. Richard and G. Uwe, 'Understanding the Link Between Social and Spatial Distance in the Crime World', in Proceedings of the 20th International Conference on Advances in Geographic Information Systems (SIGSPATIAL '12), Redondo Beach, California, 2012, pp. 550-553.
[5] S. Nath, 'Crime Pattern Detection Using Data Mining', in Web Intelligence and Intelligent Agent Technology Workshops, 2006. WI-IAT 2006 Workshops. 2006 IEEE/WIC/ACM International Conference on, 2006, pp. 41,44.
[6] Crimereports.com, 2015. [Online]. Available: https://www.crimereports.com. [Accessed: 20- May- 2015].
[7] S. Chainey, L. Tompson and S. Uhlig, 'The Utility of Hotspot Mapping for Predicting Spatial Patterns of Crime', Security Journal, vol. 21, no. 1-2, pp. 4-28, 2008.
[8] Data.denvergov.org, 'Denver Open Data Catalog: Crime', 2015. [Online]. Available: http://data.denvergov.org/dataset/city-and-county-of-denver-crime. [Accessed: 20- May- 2015].
[9] Imgh.us, 2015. [Online]. Available: http://imgh.us/neighborhood_map.jpg. [Accessed: 20- May- 2015].
[10] O. Knowledge, 'Crime — Datasets - US City Open Data Census', Us-city.census.okfn.org, 2015. [Online]. Available: http://us-city.census.okfn.org/dataset/crime-stats. [Accessed: 20- May- 2015].
[11] Laalmanac.com, 'City of Los Angeles Planning Areas Map', 2015. [Online]. Available: http://www.laalmanac.com/LA/lamap3.htm. [Accessed: 20- May- 2015].

[12] Data.denvergov.org, 'Denver Open Data Catalog: Census Neighborhood Demographics (2010)', 2015. [Online]. Available: http://data.denvergov.org/dataset/city-and-county-of-denver-census-neighborhood-demographics-2010. [Accessed: 20- May- 2015].

[13] GitHub, 'asaini/Apriori', 2015. [Online]. Available: https://github.com/asaini/Apriori. [Accessed: 20- May- 2015].

[14] Scikit-learn.org, '3.3. Model evaluation: quantifying the quality of predictions — scikit-learn 0.17.dev0 documentation', 2015. [Online]. Available:
http://scikit-learn.org/dev/modules/model_evaluation.html. [Accessed: 20- May- 2015].